\documentclass[runningheads]{llncs}
\usepackage{graphicx}
\usepackage{comment}
\usepackage{gensymb}
\usepackage{amsmath,amssymb} 
\usepackage{subfloat}
\usepackage{pifont}
\newcommand{\cmark}{\ding{51}}%
\newcommand{\xmark}{\ding{55}}%
\usepackage{color}
\def\etal{\emph{et al.}}
\makeatletter
\newcommand{\printfnsymbol}[1]{%
  \textsuperscript{\@fnsymbol{#1}}%
}
\makeatother

\usepackage{microtype}
\usepackage[pagebackref=true,breaklinks=true,letterpaper=true,colorlinks,bookmarks=false]{hyperref}
\usepackage{booktabs}

\begin{document}
\pagestyle{headings}
\mainmatter
\def\ECCVSubNumber{1483}  

\title{Sep-Stereo: Visually Guided Stereophonic Audio Generation by Associating Source Separation} 

\titlerunning{Sep-Stereo}
%
\author{Hang Zhou\thanks{Equal contribution.} \and
Xudong Xu\printfnsymbol{1} \and Dahua Lin \and Xiaogang Wang \and
Ziwei Liu}
\authorrunning{H. Zhou et al.}
%
\institute{CUHK - SenseTime Joint Lab, The Chinese University of Hong Kong 
\email{\{zhouhang@link,xx018@ie,dhlin@ie,xgwang@ee\}.cuhk.edu.hk}
\email{zwliu.hust@gmail.com}
}
\maketitle

\begin{abstract}
Stereophonic audio is an indispensable ingredient to enhance human auditory experience.
Recent research has explored the usage of visual information as guidance to generate binaural or ambisonic audio from mono ones with stereo supervision. 
However, this fully supervised paradigm suffers from an inherent drawback: the recording of stereophonic audio usually requires delicate devices that are expensive for wide accessibility.
To overcome this challenge, we propose to leverage the vastly available mono data to facilitate the generation of stereophonic audio.
Our key observation is that the task of visually indicated audio separation also maps independent audios to their corresponding visual positions, which shares a similar objective with stereophonic audio generation.
We integrate both stereo generation and source separation into a unified framework, \textbf{Sep-Stereo}, by considering source separation as a particular type of audio spatialization.
Specifically, a novel associative pyramid network architecture is carefully designed for audio-visual feature fusion. 
Extensive experiments demonstrate that our framework can improve the stereophonic audio generation results while performing accurate sound separation with a shared backbone\footnote{Code, models and demo video are available at \url{https://hangz-nju-cuhk.github.io/projects/Sep-Stereo}.}.
\end{abstract}

\section{Introduction}
\begin{figure}[t]
    \includegraphics[width=1\linewidth]{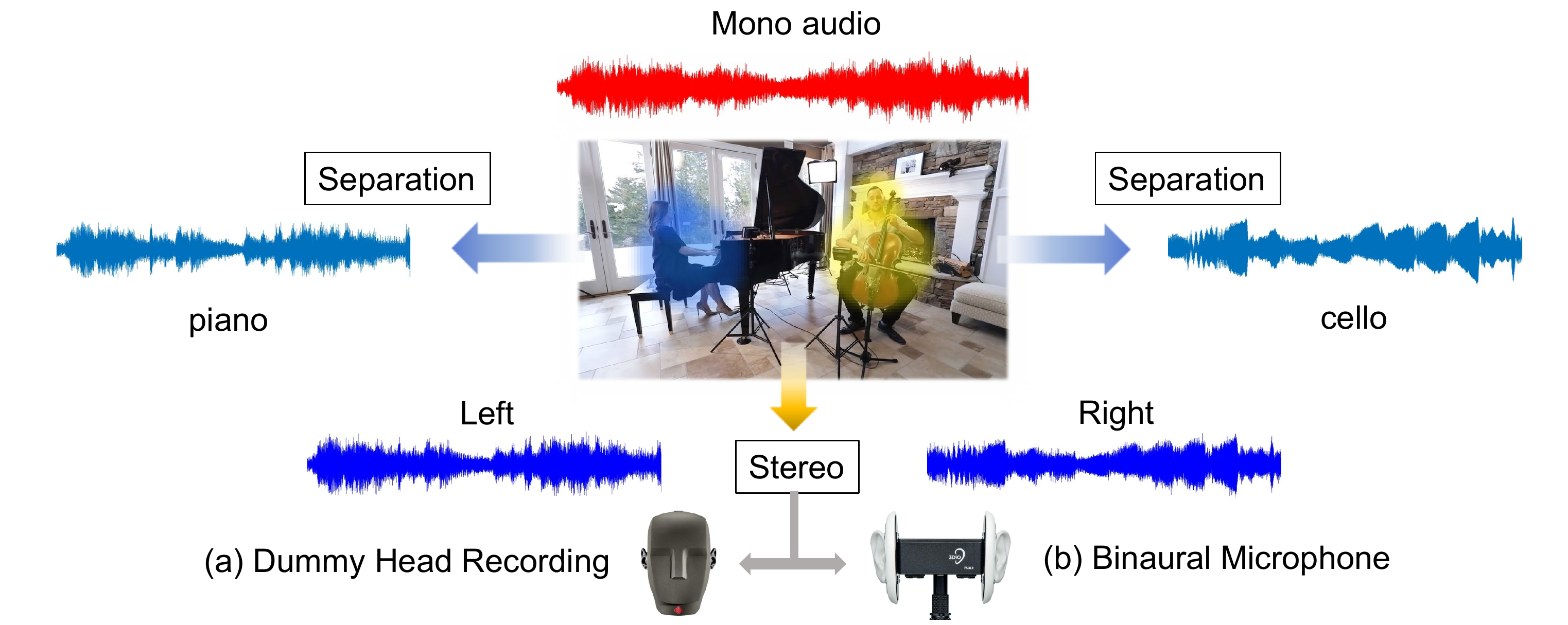}
    \caption{We propose to integrate audio source separation and stereophonic generation into one framework. Since both tasks build associations between audio sources and their corresponding visual objects in videos. Below is the equipment for recording stereo: (a) dummy head recording system, (b) Free Space XLR Binaural Microphone of 3Dio. The equipment is not only expensive but also not portable. 
    This urges the need for leveraging mono data in stereophonic audio generation
}
\label{fig:fig1}
\end{figure}
Sight and sound are both crucial components of human perceptions.
Sensory information around us is inherently multi-modal, mixed with both pixels and vocals. More importantly, the stereophonic or spatial effect of sound received by two ears gives us the superiority to roughly reconstruct the layout of the environment, which complements the spatial perception in the vision system.
This spatial perception of sound makes it appealing for content creators to create audio information with more than one channel. For example, the user's experience will be greatly promoted if stereo music instead of mono is provided when watching a recording of a concert.

However, it is still inconvenient for portable devices to record stereophonic audio. Normally, cell phones and cameras have only mono or line array microphones that can not record real binaural audio. 
To achieve such goals, dummy head recording systems or fake auricles need to be employed for creating realistic 3D audio sensations that humans truly perceive. This kind of system requires two microphones attached to artificial ears (head) for mimicking a human listener in the scenery, as shown in Fig.~\ref{fig:fig1}. Due to the cost and weight of the devices, such recorded binaural audio data is limited, particularly the ones associated with visual information. Therefore, developing a system to generate stereophonic audio automatically from visual guidance is highly desirable.


To this end, Gao and Grauman~\cite{gao20192} contribute the FAIR-Play dataset collected with binaural recording equipment. Along with this dataset, they propose a framework for recovering binaural audio with the mono input. Nevertheless, their method is built in a data-driven manner, fusing visual and audio information with a simple feature concatenation, which is hard to interpret. Moreover, the data is valuable yet insufficient to train a network that can generalize well on all scenarios. 
Similarly, Morgado \etal~\cite{morgado2018self} explore to use 360\degree~ videos uploaded on youtube for generating ambisonic audio.

On the other hand, videos recorded with a single channel are much easier to acquire. Ideally, it would be a great advantage if networks can be benefited from training with additional mono-only audios without stereophonic recordings.
This paradigm of leveraging unlabeled data has rarely been explored by previous research because of its inherent challenge.
%
%
We observe that \textbf{an essential way to generate stereo audio requires disentangling multiple audio sources and localizing them, which is similar to the task of visually indicated sound source separation}~\cite{zhao2018sound,xu2019recursive,zhao2019sound,Gan_2020_CVPR}.
While previous methods have also explored performing separation along with generating spatial audio, they either implement it as an intermediate representation by using stereo data as supervision~\cite{morgado2018self}, or train another individual network for separation~\cite{gao20192}.
Thus it is valuable to investigate a new way of leveraging both mono and stereo together for solving both problems. Moreover, it is particularly desirable if the two tasks can be solved within one unified framework with a shared backbone as illustrated in Fig.~\ref{fig:fig1}. 
However, there are considerable differences between these two tasks and great challenges to overcome. The details are explained in Section~\ref{sec:3.1}.

%
Our key insight is to \textbf{regard the problem of separating two audios as an extreme case of creating binaural audio}. More specifically, we perform duet audio separation with the hypothesis that the two sources are only visible at the edges of human sight. And no visual information can be provided about the whole scene. Based on this assumption and our observation above, we propose to explicitly associate different local visual features to different responses in spectrograms of audios. In this manner, the intensity of sound can be represented on both the image domain visually and the audio domain auditorily. A novel \textbf{associative pyramid network} architecture is proposed for better fusing the information within the two modalities. 
The whole learning process can be divided into two parts, namely \textbf{separative learning} and \textbf{stereophonic learning}. . We perform multi-task training with two different sets of data: 
mono ones (MUSIC~\cite{zhao2018sound}) for separation, and binaural ones (FAIR-Play~\cite{gao20192}) for stereophonic audio generation. Following traditional multi-task learning settings, the two learning stages share a same backbone network but learn with different heads with the same architecture. This framework is uniformly called \textbf{Sep-Stereo}.

Extensive experiments regarding stereophonic audio generation have validated the effectiveness of our proposed architecture and learning strategy. At the same time, the trained model can preserve competitive results on the task of audio separation simultaneously. Moreover, we show that with the aid of mono data, our framework is able to achieve generalization under a low data regime using only a small amount of stereo audio supervision.

Our \textbf{contributions} are summarized as follows: {\bf 1)} We unify audio source separation and stereophonic audio generation into a principled framework, \textbf{Sep-Stereo}, which performs joint \textbf{separative} and \textbf{stereophonic} learning. {\bf 2)} We propose a novel \textbf{associative pyramid network} architecture for coupling audio-visual responses, which enables effective training of both tasks simultaneously with a shared backbone. {\bf 3)} Our \textbf{Sep-Stereo} framework has a unique advantage of leveraging mono audio data into stereophonic learning. Extensive experiments demonstrate that our approach is capable of producing more realistic binaural audio while preserving satisfying source separation quality.

\section{Related Works}

\subsection{Joint Audio-Visual Learning} 

The joint learning of both audio and visual information has received growing attention in recent years~\cite{zhu2020deep,gao2020listen,gao2020visualechoes,rao2020local,huang2020movie}. 
By leveraging data within the two modalities, researchers have shown success in learning audio-visual self-supervision~\cite{aytar2016soundnet,arandjelovic2017look,arandjelovic2018objects,korbar2018cooperative,owens2018audio,Hu_2019_CVPR}, audio-visual speech recognition~\cite{Hu_2016_CVPR,Chung_2017_CVPR,zhou2019talking,9054127}, localization~\cite{zhao2018sound,senocak2018learning,rouditchenko2019self,qian2020learning}, 
event localization (parsing)~\cite{tian2018audio,wu2019dual,tian2020audio}, audio-visual navigation~\cite{gan2019look,chen2019audio}, 
cross-modality generation between the two modalities~\cite{chung2017you,visual2sound,chen2017deep,chen2018lip,zhou2019talking,chen2019hierarchical,zhu2018high,zhou2019vision,NIPS2019_8768,zhou2020makeittalk} and so on.
General representation learning across the two modalities is normally conducted in a self-supervised manner.
Relja \etal ~\cite{arandjelovic2017look,arandjelovic2018objects}
propose to learn the association between visual objects and sound, which supports localizing the objects that sound in an image. 
Owens \etal ~\cite{owens2018audio} and Korbar \etal ~\cite{korbar2018cooperative} train neural networks to predict whether video frames and audios are temporally aligned. 
%
Researchers have also explored the possibility of directly generating sound according to videos~\cite{owens2016visually,chen2017deep,visual2sound},
which is more related to our task. 
%
Different from their aims, our Sep-Stereo framework exploits visual-audio correspondence to improve the generation of stereophonic audios.

\subsection{Audio Source Separation and Spatialization} 

\noindent{\textbf{Source Separation.}} Source separation with visual guidance has been an interest of research for decades~\cite{fisher2001learning,maganti2007speech,parekh2017motion,gao2018learning,zhao2018sound,zhao2019sound,afouras2018conversation,xu2019recursive,ephrat2018looking,Gan_2020_CVPR}. Compared with audio-only source separation, visual information could provide rich clues about the types and movements of audio sources. Thus the performance of audio separation is expected to improve with the guidance of vision.
Recently, deep learning has been widely applied into this filed of research. For separating speech segments, Afouras \etal ~\cite{afouras2018conversation} propose to leverage mouth movements as guidance, and Ephrat \etal ~\cite{ephrat2018looking} use cropped human faces. Owens \etal ~\cite{owens2018audio} do not crop faces and modify their pipeline from learning synchronization. On the other hand, instrumental music~ \cite{zhao2018sound,zhao2019sound,gao2018learning,gao2019co,xu2019recursive} is the field that we care more about. 
Gao \etal ~\cite{gao2018learning} propose to combine non-negative matrix factorization (NMF) with audio features. Zhao \etal ~\cite{zhao2018sound} use a learnable U-Net instead, and match feature activations with different audio channels. Based on this work,  motion information is merged into the main framework to achieve better performance in \cite{zhao2019sound}. In \cite{gao2019co}, object detection and instrument labels are leveraged to co-separating sound. 
In our work, we will not model motion explicitly, thus adopt a similar setting as \cite{zhao2018sound}.

\noindent{\textbf{Spatialization.}}
Visually guided audio spatialization has received relatively less attention~\cite{li2018scene,lu2019self,morgado2018self,gao20192,gan2019self}  compared with separation. Recently, Li \etal ~\cite{li2018scene} leverage synthesised early reverberation and measured late reverberation tail to generate stereo sound in a specific room, which cannot generalize well to other scenarios.
With the assistance of deep learning, Morgado \etal ~\cite{morgado2018self} propose to generate ambisonic for 360\degree ~videos using recorded data as self-supervision. The work mostly related to ours is~\cite{gao20192}. They contribute a self-collected binaural audio dataset, and propose a U-Net based framework for mono-to-binaural generation on normal field of view (NFOV) videos. These works all only leverage the limited stereophonic audio. In this paper, we propose to boost the spatialization performance with additional mono data.

\section{Our Approach}
 Our proposed framework, \textbf{Sep-Stereo}, is illustrated in Fig.~\ref{fig:pipeline}. This whole pipeline consists of two parts: (a) stereophonic learning and (b) separative learning. We will first introduce the overall framework of visually guided stereophonic audio generation and source separation (section~\ref{sec:3.1}). Then we demonstrate how our proposed network architectures can effectively associate and integrate audio and visual features into a unified network with a shared backbone. 

\subsection{Framework Overview}

\noindent{\textbf{Stereophonic Learning.}}
The whole process of stereophonic learning is depicted in the lower part in Fig.~\ref{fig:pipeline}. 
In the setting of stereophonic learning, we care for the scenario that human perceives and has access to binaural data. The visual information $V_s$ corresponds to its audio recording of the left ear $a_l(t)$ and the right-ear one $a_r(t)$. Notably, all spatial information is lost when they are averaged to be a mono clip $a_{mono} = (a_l + a_r)/2$, and our goal is to recover left and right given the mono and video. 
We operate in the Time-Frequency (TF) domain by transferring audio to spectrum using Short-Time Fourier Transformation (STFT) as a common practice.
Here we use $S^t_l$ and $S^t_r$ to denote the STFT of the ground truth left and right channels, with $t$ here represents ``target''. 
The input of our network is the mono audio by averaging the STFT of the two audio channels:
\begin{align}
\label{eq:2}
    S_{mono} = (S^t_l + S^t_r) / 2 = \text{STFT}(a_{mono}).
\end{align}
This can be verified by the property of the Fourier Transformation.
Please note that due to the complex operation of STFT, each spectrum $S = S_R + j*S_I$ is a complex matrix that consists of the real $S_R$ and imagery part $S_I$. So the input size of our audio network is $[T, F, 2]$ by stacking the real and imagery channels.

\noindent{\textbf{Separative Learning.}}
The task of separation is integrated for its ability to leverage mono data.
Our separative learning follows the Mix-and-Separate training procedure~\cite{zhao2018sound}, where we elaborately mix two independent audios as input and manage to separate them using ground truth as supervision. It is illustrated at the top of Fig.~\ref{fig:pipeline}.
Given two videos $V_A$ and $V_B$ with only mono audios accompanied, the input of the separative phase is the mixture of two mono audios $a_{mix} = (a_{A} + a_{B}) / 2$. They can be represented in the STFT spectrum domain as $S^t_A$, $S^t_B$ and $S_{mix}$. 
Aiming at disentangling the mixed audio, separative learning targets at recovering two mono audios with the guidance of corresponding videos.

\label{sec:3.1}
\noindent{\textbf{Connections and Challenges.}} Apart from our observation that both tasks connect salient image positions with specific audio sources, they all take mono audio clips as input and attempt to split them into two channels. One can easily find a mapping from stereo to separation as: $\{ S_{mono} \Rightarrow S_{mix}$, $S^t_l  \Rightarrow S^t_A$, $S^t_r \Rightarrow S^t_B\}$.  From this point of view, the two tasks are substantially similar.
However, the goals of the two tasks are inherently different. While each separated channel should contain the sound of one specific instrument, both sources should be audible, \textit{e.g.} in the task of stereo for a scene shown in Fig.~\ref{fig:fig1}. Also, the spatial effect would exist if there is only one source, but separation is not needed in such a case. As for the usage of visual information, the separation task aims at finding the most salient area correctly while the stereo one is affected by not only the sources' positions but also the environment's layout.
Thus neither an existing stereo framework~\cite{gao20192} nor a separation one~\cite{zhao2018sound} is capable of handling both tasks.

\subsection{Associative Neural Architecture}

\begin{figure*}[t!]
\centering
\includegraphics[width=1\linewidth]{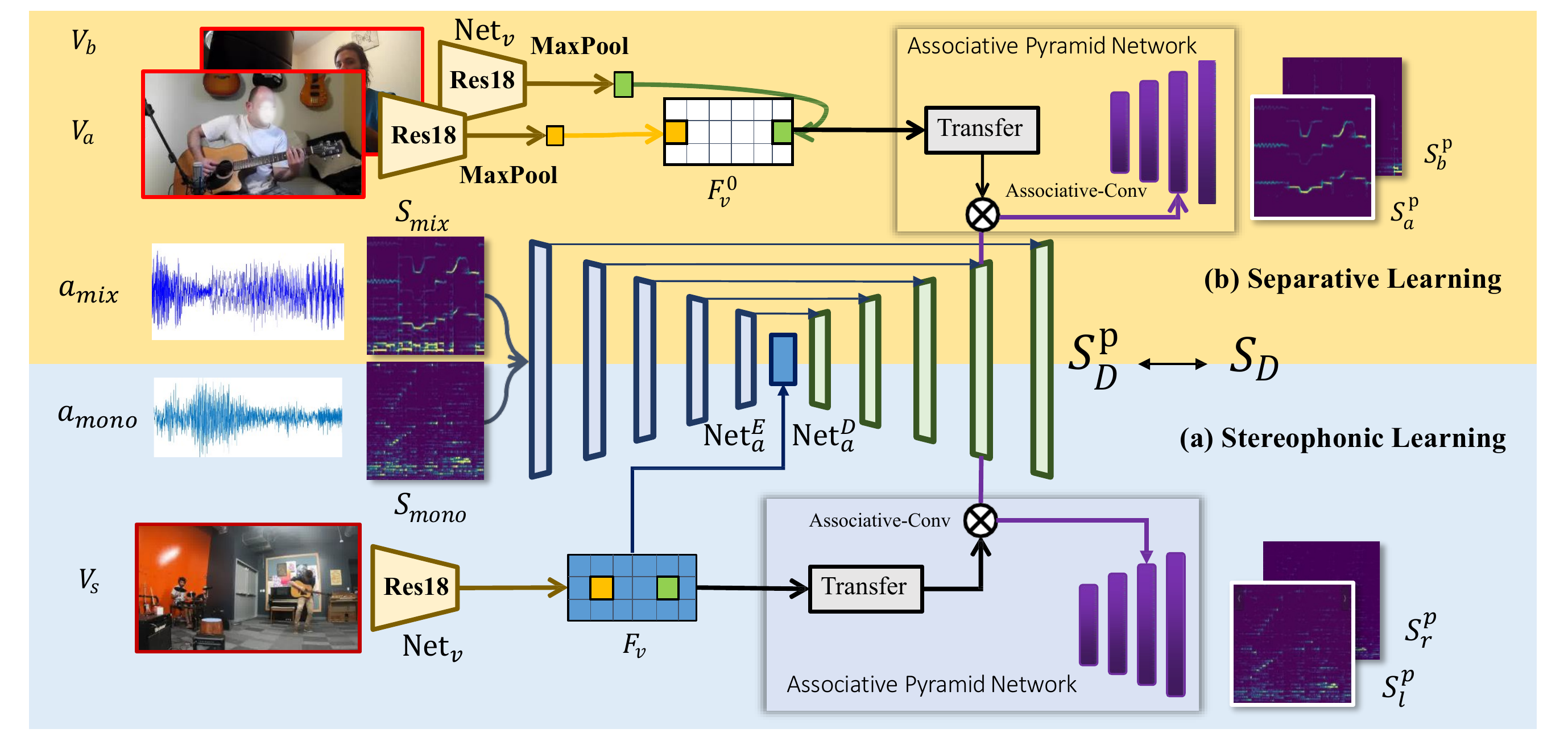}
\caption{\textbf{
The whole pipeline of our Sep-Stereo framework.} It aims at learning the associations between visual activation and the audio responses. The framework consists of a U-Net backbone whose weights are shared, and our proposed Associative Pyramid Network (APNet). The visual features are firstly extracted and fused with audio features through a multi-scale feature mapping network (APNet). To perform multitask training, (a) stereophonic learning and (b) separative learning stages are leveraged to tackle this problem. One set of APNet parameters are trained for each task}
\label{fig:pipeline}
\end{figure*}


\noindent{\textbf{Backbone Network.}}
Our audio model is built upon Mono2Binaural~\cite{gao20192}, which is a conditional U-Net~\cite{ronneberger2015u}. This audio backbone is denoted as $\text{Net}_a$. It consists of skip-connected encoder $\text{Net}^E_a$ and decoder $\text{Net}^D_a$. 
The visual features are extracted using a visual encoder with ResNet18~\cite{he2016deep} architecture called $\text{Net}_v$. The input video clip with input dimension $ [T, W_v, H_v, 3]$ is encoded into a feature map $F_v$ of size $[w_v, h_v, C_v] $,  by conducting a temporal max pooling operation. We assume that the feature map would correspond to each spatial part in the original video with high responses on salient positions. 

\begin{figure}[t]
    \centering
    \includegraphics[width=1\linewidth]{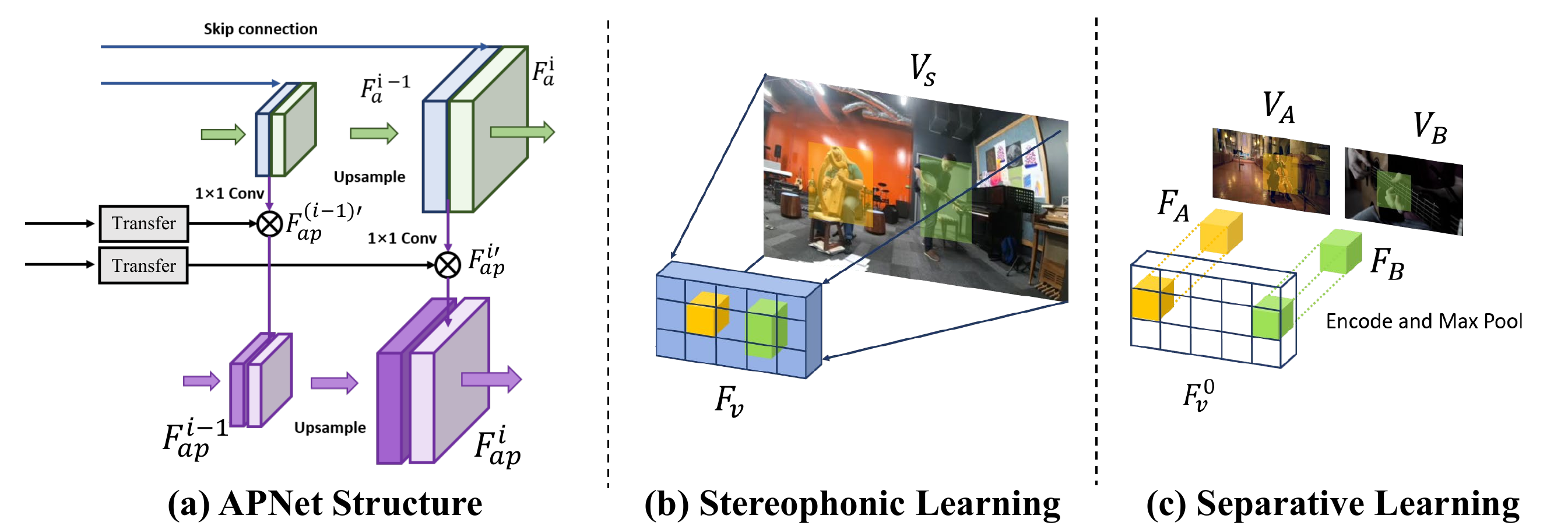}
    \caption{Figure (a) is the architecture of the Associative Pyramid Network. After kernel transfer of the visual feature, it associates audio channels with spatial visual information through $1 \times 1$ convolution and coarse-to-fine tuning. Figure (b) and (c) shows the difference between stereo and separative learning with respect to visual embedding. For (b) stereophonic learning, a visual feature map $F_v$ is directly extracted, with each bin corresponds to a region in the image. While in (c), the visual feature is max-pooled to a $1 \times 1$ bin and manually sent to an all-zero feature map $F^0_v$. This process is the \emph{rearrangement} of visual feature
}
\label{fig:apn}
\end{figure}

\noindent{\textbf{Associative Pyramid Network.}}
Based on the backbone network, we propose a novel Associative Pyramid Network (APNet) for both learning stages. It is inspired by PixelPlayer~\cite{zhao2018sound} that maps one vision activation with one source feature map, but with a different formulation and underlying motivation. Our key idea is to associate different intensities of audio sources with different vision activations in the whole scene with feature map re-scheduling. As illustrated in Fig.~\ref{fig:pipeline} and \ref{fig:apn}, it works as a side-way network along-side the backbone in a coarse-to-fine manner.

We operate on each layer of the decoder $\text{Net}^D_a$ in the U-Net after the upsample deconvolutions. Suppose the $i$th deconv layer's feature map $F^i_a$ is of shape [$W^i_a,  H^i_a, C^i_a$], we first reshape $F_v$ to [$(w_v \times h_v), C_v$] and multiply it by a learned weight with size [$C_v, C^i_a$] to be $K^i_v$ with dimension [$1, 1, C^i_a,  (h_v \times w_v)$]. This is called the kernel transfer operation.
Then $K^i_v$ operates as a $1 \times 1$ 2D-convolution kernel on the audio feature map $F^i_a$, and renders an entangle audio-visual feature  $F^{i'}_{ap}$ of size $[W^i_a,  H^i_a, (h_v \times w_v)]$ . This process can be formulated as:
\begin{align}
\label{eq:3}
    F^{i'}_{ap} = \underset{K^i_v}{\text{Conv2d}}
    ({F^i_a}).
\end{align}
Note that each $[1, 1, C^i_a]$ sub-kernel of $K^i_v$ corresponds to one area in the whole image space. So basically, audio channels and positional vision information are associated through this learned convolution. This operation is named Associative-Conv.
The output feature $F^{i'}_{ap}$ can be regarded as the stack of audio features associated with different visual positions.

The $F^{1'}_{ap}$ is the first layer of APNet $F^{1}_{ap}$. When $i > 1$, the ($i - 1$)th feature map $F^{i - 1}_{ap}$ will be upsampled through a deconvolution operation to be of the same size as $F^{i'}_{ap}$. Then a new feature map $F^{i}_{ap}$ can be generated by a concatenation:
\begin{align}
\label{eq:3}
    F^{i}_{ap} = \text{Cat}([\text{DeConv}(F^{i - 1}_{ap}), F^{i'}_{ap}]).
\end{align}
In this way, the side-way APNet can take advantage of the pyramid structure by coarse-to-fine tuning with both low-level and high-level information. 

The goal of APNet is to predict the target left and right channels' spectrum in both learning stages. Thus, two parallel final convolutions are applied to the last layer of $F^{i}_{ap}$, and map it to two outputs with real and imagery channels. As discussed before, each channel in the APNet is specifically associated with one visual position, the final convolution acts also as a reconfiguration to different source intensities. 

\subsection{Learning Sep-Stereo}

Directly predicting spectrum is difficult due to the large dynamic range of STFTs, so we predict the complex masks $M = \{M_R, M_I\}$ following~\cite{gao20192} as our objectives. Suppose the input spectrum is $S_{mono} = S_{R(mono)} + j*S_{I(mono)}$, then a prediction can be written as:
\begin{align}
\label{eq:5}
    S^p = (S_{R(mono)} + j*S_{I(mono)})(M_R + j*M_I).
\end{align}
The outputs of our networks are all in the form of complex masks, and the predictions are made by the complex multiplication stated above.

\noindent\textbf{Stereophonic Learning.}
The base training objective of the backbone network for stereo is to predict the subtraction of the two spectrums
$S^t_D = (S^t_l - S^t_r) / 2$ as proposed in~\cite{gao20192}.
The left and right ground truth can be written as:
\begin{align}
\label{eq:4}
    S^t_r = S_{mono} + S^t_D,~~S^t_l = S_{mono} - S^t_D.
\end{align} 
The backbone network is to predict the difference spectrum $S^p_D$, the training objective is:
\begin{align}
\label{eq:7}
    L_D = || S^t_D - S^p_D ||^2_2.
\end{align}
The APNet's outputs are the left and right spectrums, the loss fuction is:
\begin{align}
\label{eq:8}
    L_{rl} = || S^t_l - S^p_l ||^2_2 + || S^t_r - S^p_r ||^2_2.
\end{align}

\noindent\textbf{Separative Learning.}
While the backbone network and APNet seem to be suitable for learning binaural audio, it has the advantage of handling separation by manually modifying features. Our key insight is to regard separation as an extreme case of stereo, that the visual information is only available at the left and right edges. Normally, the visual feature map  $F_v$ is a global response which contains salient and non-salient regions. During the separation stage, we manually create the feature map. 

Specifically, we adopt max-pooling to the visual feature map $F_v$ to be of size $[1, 1, C_v]$. The feature vectors for video $A$ and $B$ are denoted as $F_A$ and $F_B$ respectively. Then we create an all-zero feature map $F^0_v$ of the same size as $[w_v, h_v, C_v]$ to serve as a dummy visual map. Then the max-pooled vectors are sent to the left and right most positions as illustrated in Fig.~\ref{fig:apn}. It can be written as:
\begin{align}
\label{eq:9}
    F^0_v(\lceil H/2 \rceil,~1) = F_A,~~F^0_v(\lceil H/2 \rceil, W) = F_B.
\end{align}
Then we replace the $F_v$ with $F^0_v$. This process is called the \textbf{rearrangement} for visual feature.



The intuition for the separative learning to work is based on our design that each channel of the APNet layers corresponds to one visual position. While with separative learning, take the left-ear for instance. Most information correlates to the left-ear is zero, thus the left-ear spectrum should correspond to only the left-most visual information.
Training the separation task provides especially the backbone network with more audio spectrum and vision information instead of only overfitting to the limited binaural data. 
Besides, it is also assumed that the non-salient visual features also help our APNet understand the sound field. Thus, without the environment information, we can expect the network to implicitly ignore the distribution of sound around the space but focus on the two sides.

At the separative learning stage, only the APNet predicts the masks for audios $A$ and $B$, as the left and right channels in the same way as Eq.~\ref{eq:5}. The predicted spectrums can be represented as $S^p_a$ and $S^p_b$. So the training objective is:
\begin{align}
\label{eq:10}
    L_{ab} = || S^t_a - S^p_a ||^2_2 + || S^t_b - S^p_b ||^2_2.
\end{align}

\noindent{\textbf{Final Objective.}} The final objective of our network is the combination of all the losses for training stereo and separation.
\begin{align}
\label{eq:11}
    L_{all} = L_{D} + {\lambda_1}L_{rl} + {\lambda_2}L_{ab}
\end{align}
where $\lambda_1$ and $\lambda_2$ are loss weights that are empirically set to 1 in the experiments through cross-validation.

\section{Experiments}

\subsection{Implementation Details}

\noindent\textbf{Preprocessing.} 
%
We fix all of our audio sampling rate to 16kHz and clip the raw audios to ensure their values are between -1 and 1. For performing STFT, our window size is 512, and the hop length is 160. During stereophonic training, we sample a 0.63s clip randomly from the whole 10s video. Thus can lead to an STFT map with the size of [$257 , 64$]. 
Separative learning samples a 0.63s clip from each individual video as well, and mixes them up as inputs. Other configurations are the same as \cite{gao20192}. The length of the sliding window for testing is 0.1s.
The videos are extracted to frames at 10 fps. 
At each training time step, the center frame is used as the input of the visual embedding network.

\noindent\textbf{Model Configurations.} 
The backbone audio U-Net $\text{Net}_a$ is borrowed from~\cite{gao20192}, which consists of 5 downsample convolution and 5 de-convolution layers with 4 skip connections between feature maps of the same scale. The Associative Pyramid Network consists of 4 Associative-Conv which couples visual features with audio features. Additionally, there are 3 upsampling operations in APNet. 
The visual embedding network $\text{Net}_v$ is adopted from~\cite{zhao2018sound},
which is a modified ResNet18 network~\cite{he2016deep}. The final pooling and fully-connected layers are removed from this network, and the dilation of the network's kernels is 2. 
Thus $F_v$ is of size $[14, 7, 512]$, where 512 is its channel size.

\noindent\textbf{Training Details.}
%
The networks are trained using Adam~\cite{kingma2014adam} optimizer with learning rate at 5e-4 and batch size 144. 
For stereophonic learning, we use the same data augmentation diagram as Mono2Binaural~\cite{gao20192}.
For separative learning, the amplitude of selected audio is augmented with a random scale disturb of 0.5 to 1.5. The separative learning part is firstly trained, then both data of stereo and separation are sent into the network at the same time. 
Our original design is to share the backbone and APNet parameters through both learning stages. However, it requires careful tuning for both tasks to converge simultaneously. In our final version, the parameters of $\text{Net}_a$ and  $\text{Net}_v$ are \textbf{shared} across two learning stages while \textbf{different} sets of APNet parameters are trained for different stages. 
As the the backbone takes up most of the parameters, sharing it with separative learning is extremely important for stereophonic learning. Moreover, visual information is also fused into the backbone, thus our insights all stand even without sharing the APNet parameters.

\subsection{Datasets and Evaluation Metrics}
In the sense of improving audiences' experiences, videos with instrumental music are the most desired scenario for stereophonic audio. Thus in this paper, we choose music-related videos and audios as a touchstone. 
Our approach is trained and evaluated on the following datasets:

\noindent\textbf{FAIR-Play.} The FAIR-Play dataset is proposed by Gao and Grauman~\cite{gao20192}. 
It consists of 1,871 video clips. The train/val/test has already been split by the authors. We follow the same split and evaluation protocol of \cite{gao20192}.

\noindent\textbf{YT-MUSIC.} This is also a stereophonic audio dataset that contains video recordings of music performances in $360\degree$ view. It is the most challenging dataset collected in paper~\cite{morgado2018self}. As the audios are recorded in first-order ambisonics, we convert them into binaural audios using an existing converter and also follow the protocol of \cite{gao20192}.

\noindent\textbf{MUSIC.} We train and evaluate the visually indicated audio source separation on the solo part of MUSIC dataset~\cite{zhao2018sound}. Note that in our comparing paper~\cite{zhao2019sound}, this dataset is enriched to a version with 256 videos for testing named MUSIC21. We follow this setting and use an enriched version with 245 videos for testing, so the comparisons are basically fair. Please be noted that our whole Sep-Stereo model with separative learning is trained on this dataset.

\noindent\textbf{Stereo Evaluation Metrics.}
We evaluate the performance of audio spatialization using similar metrics used in Mono2Binaural~\cite{gao20192} and Ambisonics~\cite{morgado2018self}. 
\begin{itemize}
\item  \textit{STFT Distance ($\text{STFT}_D$).} As all existing methods are trained in the form of STFT spectrum, it is natural to evaluate directly using the training objective on the test set.
\item  \textit{Envelope Distance ($\text{ENV}_D$).} As for evaluations on raw audios, we use the envelope distance. It is well-known that direct comparison on raw audios is not informative enough due to the high-frequency nature of audio signals. So we follow~\cite{morgado2018self} to use differences between audio envelopes as a measurement. 
\end{itemize}

\noindent\textbf{Separation Evaluation Metrics.} We use these source separation metrics following~\cite{zhao2018sound}: Signal-to-Distortion Ratio (SDR), Signal-to-Interference  Ratio  (SIR),  and  Signal-to-Artifact Ratio (SAR). The units are dB.



\setlength{\tabcolsep}{6.5pt}
\begin{table*}[t] 
\begin{center}
\caption{Comparisons between different approaches on FAIR-Play and YT-Music dataset with the evaluation metric of STFT distance and envelop distance. The lower the score the better the results. The training data types for each method are also listed. It can be seen from the results that each component contributes to the network
}
\label{table:exp1}
\begin{tabular}{lcccccc}
\toprule
 & \multicolumn{2}{c}{Training Data}& \multicolumn{2}{c}{FAIR-Play }& \multicolumn{2}{c}{YT-Music}\\

Method &Stereo &Mono & $\text{STFT}_D$ & $\text{ENV}_D$ &$\text{STFT}_D$& $\text{ENV}_D$\\

\midrule
Mono2Binaural~\cite{gao20192}  &\cmark &\xmark & 0.959 & 0.141 & 1.346 & 0.179  \\
Baseline (MUSIC) &\cmark &\cmark & 0.930  & 0.139 &  1.308 & 0.175\\
Assoicative-Conv &\cmark &\xmark & 0.893  & 0.137&  1.147 & 0.150\\
APNet &\cmark &\xmark & 0.889 & 0.136&  1.070 & 0.148\\
\textbf{Sep-Stereo (Ours)}  &\cmark &\cmark& \textbf{0.879}  & \textbf{0.135} &\textbf{1.051} & \textbf{0.145}\\

\bottomrule
\end{tabular}
\end{center}
\end{table*}
\setlength{\tabcolsep}{2pt}



\subsection{Evaluation Results on Stereo Generation}

As the model of Mono2Binaural is also the baseline of our model, we re-produce Mono2Binaural with our preprocessing and train it carefully using the authors' released code. 
Besides, we perform extensive ablation studies on the effect of each component in our framework brings on the FAIR-Play and MUSIC dataset. Our modification to the original Mono2Binaural are basically the following modules:

\noindent\textbf{1) Associative-Conv.} While our APNet associates the visual and audio features at multiple scales, we conduct an additional experiment with Associative-Conv operating at only the outmost layer. This module aims to validate the effectiveness of the associative operation.

\noindent\textbf{2) APNet.} Then we perform the whole process of stereo learning with the complete version of APNet. Four layers of associative mappings are utilized to perform the coarse-to-fine tuning of the stereo learning.

\noindent\textbf{3) Baseline (MUSIC).} 
It is not possible for Mono2Binaural to use mono data in its original setting. Nevertheless, we manage to integrate our separative learning into the baseline by using our \textbf{rearrangement} module for the visual feature illustrated in Fig.~\ref{fig:apn} (c). The other parts of the network remain the same. This model can also validate the advantage of our proposed separative learning over Mono2Binaural.

\noindent\textbf{4) Sep-Stereo (Ours).} Finally, we add the data from MUSIC dataset for training separation. In this model, the separative learning and stereo learning are working together towards more accurate stereophonic audio generation. 
%


The results of the experiments tested on FAIR-Play and YT-MUSIC dataset are listed in Table~\ref{table:exp1}. The ``Training Data" column shows whether these models are trained on MUSIC. Due to different preprocessing and sliding window sizes for testing, the results reported in paper~\cite{gao20192} in not directly comparable. So we use our re-produced results for comparison. It can be seen that step by step adding our module can lead to better stereo results. Particularly, adding Associative-Conv can shorten the STFT distance by a large margin, which proves that this procedure can efficiently merge audio and visual data. Then improvements can be seen when expanding it to be APNet. Finally, integrating separative learning into our framework gives the network more diverse data for training which leads to be the best outcome.

\setlength{\tabcolsep}{3pt}
\begin{table}[t] 
\caption{Source separation results on MUSIC dataset. The units are dB. $^{\dagger}$Note that DDT results are directly borrowed from the original paper~\cite{zhao2019sound}. It uses additional motion information, while other methods only use static input}
\begin{center}

\begin{tabular}{cccccc}

\toprule

 Metric & Baseline(MUSIC)& Associative-Conv &  PixelPlayer & Sep-Stereo(Ours) & DDT$^{\dagger}$ \\
\noalign{\smallskip}
\midrule

SDR & 5.21 & 5.79 & 7.67 & 8.07 & 8.29\\

SIR & 6.44 & 6.87& 14.81 & 10.14 & 14.82\\
SAR & 14.44 & 14.49 & 11.24 &15.51 &14.47\\
\bottomrule
\label{table:music}
\end{tabular}

\end{center}
\end{table}
\setlength{\tabcolsep}{1.4pt}

\begin{figure*}[t]
    \includegraphics[width=1\linewidth]{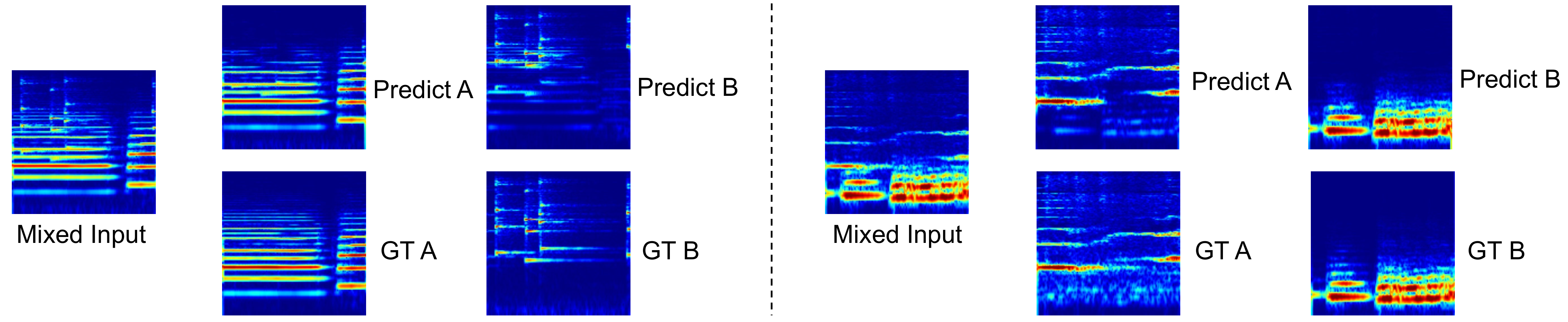}
    \caption{The visualization of separation results from spectrums. Our results are very similar to the ground truth (GT)}
\label{fig:sep}
\end{figure*}


\subsection{Evaluation Results on Source Separation} 

Our competing methods are the baseline (MUSIC) trained directly for separation, ablation of Associative-Conv, the results of self-implemented PixelPlayer~\cite{zhao2018sound} and DDT~\cite{zhao2019sound} which are originally designed for the separation task. With or without training on FAIR-Play for predicting stereo has little influence on our separation results, so we report the duet trained ones.

As shown in Table~\ref{table:music} that the baseline (MUSIC) and Associative-Conv model cannot achieve satisfying results. However, our Sep-Stereo can outperform PixelPlayer by two metrics and can keep competitive results with DDT. Note that the results of DDT are the reported ones from the original paper~\cite{zhao2019sound}, thus the results are not directly comparable. The state-of-the-art DDT uses motion information which we do not leverage. 
Reaching such a result shows the effectiveness of APNet and the value of our model. There is no doubt that we have the potential for further improvements.
We visualize two cases of separation results with duet music in Fig.~\ref{fig:sep}. It can be observed that our method can mostly disentangle the two individual spectrums from the mixed one. 

\subsection{Further Analysis}

%

\begin{figure*}[t]
    \includegraphics[width=1\linewidth]{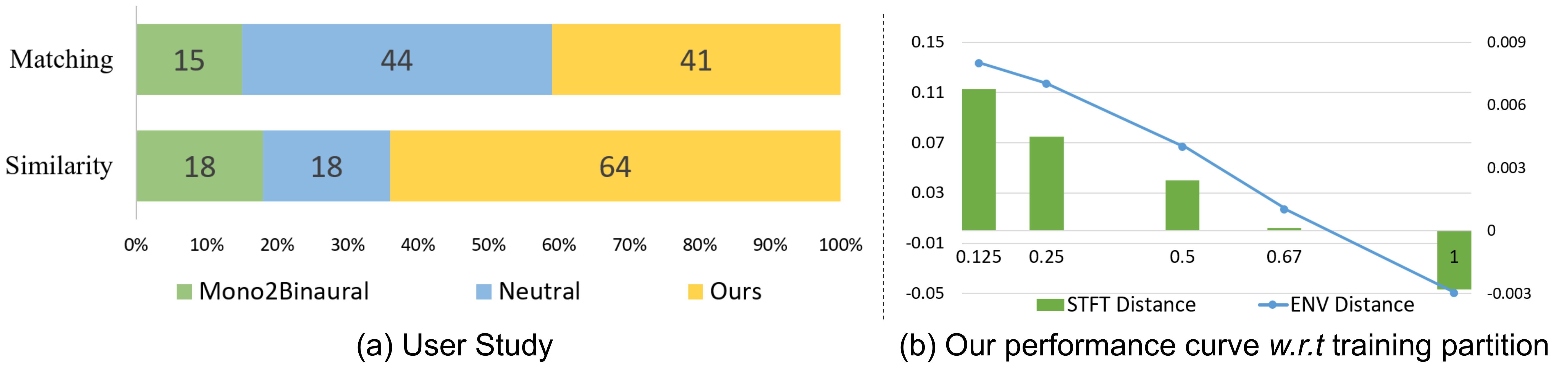}
    \caption{(a) User study for their preferences between ours and Mono2Binaural~\cite{gao20192}. The \textbf{matching} stands for audio-visual matching quality and \textbf{similarity} represents their similarity with ground truth. The \textbf{Neutral} selection means the users cannot tell which one is better. The results are shown in percentage (\%). It can be seen that ours are more preferred by users. %
    (b) The curve of our relative performances \textit{w.r.t} the percentage of training data we use on FAIR-Play. The X-axis is the fraction of training data used. Axis on the left represents STFT distance and right is ENV. The lower the better. The curve is drawn in a relative setting, where the performance of Mono2Binaural serves as the reference (zero in both metrics). It can be observed that our framework can reach their full performance using only 67\% of the training data}
\label{fig:user}
\end{figure*}
\noindent\textbf{User Study.}
We conduct user studies to verify the stereophonic quality of our results. We show the users a total of 10 cases from the FAIR-Play~\cite{gao20192} dataset. Four are results selected from Mono2Binaural's video and six are results generated by our own implementation. A total number of 15 users with normal hearing are asked to participate in the study, and a monitoring-level earphone Shure SE846 is used for conducting the study in a quiet place.

The users are asked to listen to the audio and watch the video frames at the same time. They will listen to the generated audios first and listen to the ground truth. They are responsible for telling their preferences over (1) the audio-visual \textbf{matching} quality; which of the two audios better matches the video. And (2) \textbf{similarity} to the ground truth; which of the two audios are closer to the ground truth. The users can listen to the clips multiple times. One \emph{Neutral} option is provided if it is really difficult to tell the difference.  
The results show the users' preferences in Fig.~\ref{fig:user} (a). The final results are averaged per video and per user. The table shows the ratio between selections. It can be seen that it is hard to tell the differences for certain untrained users without the ground truth. However, more people prefer our results than Mono2Binaural under both the two evaluations. The confusion is less when the ground truth is given. It can be inferred that our results are more similar to the ground truth.

\noindent\textbf{Audio-Based Visual Localization.}
We illustrate the visually salient areas learned from our model in Fig.~\ref{fig:result}. The way is to filter intense responses in feature $F_v$ back to the image space.
It can be seen that our network focuses mostly on instruments and humans, which are undoubtedly the potential sound sources. 

\noindent\textbf{Generalization under Low Data Regime.} 
We show the curve of our relative performance gains \textit{w.r.t} the percentage of training data used on FAIR-Play in Fig.~\ref{fig:user} (b). 
The curve is drawn in a relative setting, where the performance of Mono2Binaural serves as the reference (zero in both metrics). It can be observed that our framework can reach their full performance by using only 67\% of the training data, which is an inherent advantage that our separative and stereophonic learning with mono data brings under low data regime~\cite{liu2019large}.

We also highlight our model's ability of generalization to unseen scenarios by leveraging separative learning. Previous methods such as Mono2Binaural can only be trained with stereo data.
It is difficult for them to handle out-of-distribution data if no supervision can be provided. %
While our method is naturally trained on mono ones, 
by additional training on only a small portion of stereophonic data with supervision, our method can generalize to in-the-wild mono scenarios. %
The video results and comparisons can be found at \url{https://hangz-nju-cuhk.github.io/projects/Sep-Stereo}.
\begin{figure*}[t]
    \includegraphics[width=1\linewidth]{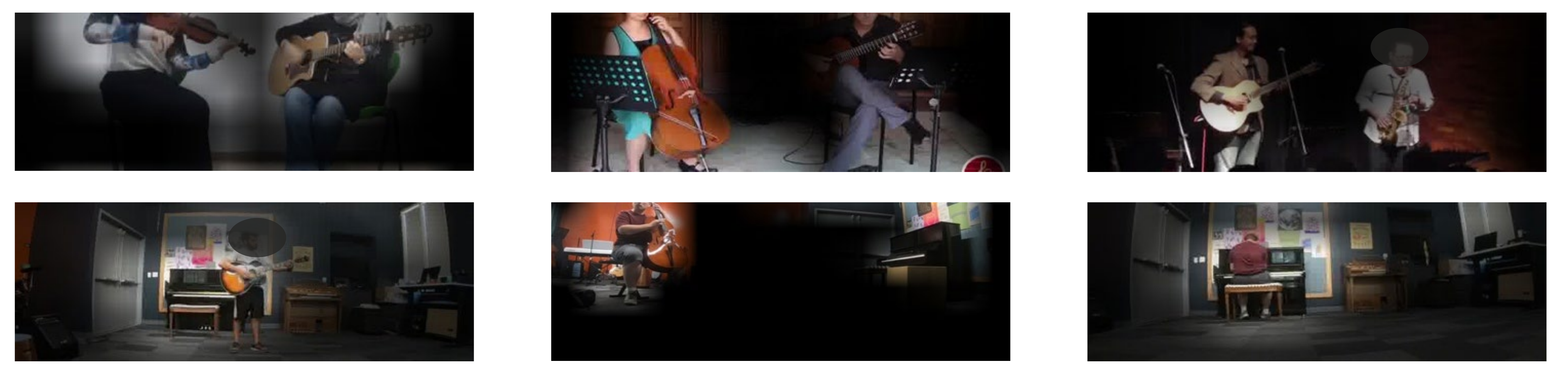}
    \caption{The visualization of the visual responses according to audio-visual associative learning. The bright places are the salient regions in the feature map, which correspond to intense audio information. The images are selected from MUSIC and FAIR-Play}
\label{fig:result}
\end{figure*}


\section{Conclusion}

In this work, we propose to integrate the task of stereophonic audio generation and audio source separation into a unified framework namely \textbf{Sep-Stereo}. We introduce a novel perspective of regarding separation as a particular type of stereo audio generation problem through manual manipulation on visual feature maps. 
We further design Associative Pyramid Network (APNet) which associates the visual features and the audio features with a learned Associative-Conv operation. 
Our proposed Sep-Stereo has the following appealing properties that are rarely achieved before: \textbf{1)} Rich mono audio clips can be leveraged to assist the learning of binaural audios. 
\textbf{2)} The task of audio separation and spatialization can be solved with a shared backbone with different heads, thus additional parameters for an entire extra network can be removed. 
\textbf{3)} Stereophonic generation can be generalized to low data regime with the aid of mono data. Extensive evaluation, analysis and visualization demonstrate the effectiveness of our proposed framework. \\

\noindent\textbf{Acknowledgements.}
%
This work is supported by SenseTime Group Limited, the General Research Fund through the Research Grants Council of Hong Kong under Grants CUHK14202217, CUHK14203118, CUHK14205615, CUHK14207814, CUHK14208619, CUHK14213616, CUHK14203518, and Research Impact Fund R5001-18.

\clearpage
%
%
\bibliographystyle{splncs04}
\bibliography{egbib}
\clearpage
\appendix
\addcontentsline{toc}{section}{Appendices}

\section*{Appendices}

\section{Demo Video Settings}

This section describes the settings of the demo video for this paper. The video can be found at \url{https://hangz-nju-cuhk.github.io/projects/Sep-Stereo}.

\subsection{Stereophonic Audio Generation}

\subsubsection{Experiments on FAIR-Play}

In our video, we first showcase of our stereophonic audio generation results on the standard FAIR-Play~\cite{gao20192} dataset. In the first three cases, we show the results of ours compared with that of Mono2Binaural~\cite{gao20192}, by using directly the videos reported by their paper. We find the validation sets these videos lying in and use the corresponding models for generation. Although the difference is not significant, one can still find that our method can create examples with more precise directional information than theirs.

\subsubsection{Experiments on MUSIC}

We then attempt to generate stereophonic audios for both in-the-wild data and synthetic duet data on MUSIC~\cite{zhao2018sound} dataset. Please be noted that all audio samples we use in the MUSIC dataset are converted to mono, thus no stereophonic training is applicable on this dataset. The results of the three models are shown:
\begin{itemize}

\item \textbf{Mono2Bianural}~\cite{gao20192}. We use this baseline model trained on FAIR-Play. Notice that this method cannot be trained with the mono data in MUSIC inherently.

\item \textbf{Sep-Stereo (Ours)}. Then is our whole Sep-Stereo model that is trained on the mono part of MUSIC and only 50\% of the data on split 1 of FAIR-Play. We find that adding more data has subtle effects on human perceptions.

\item \textbf{Unsupervised Sep-Stereo}. This is our model with only separative and no stereophonic learning. In other words, we train our model only on the mono data of MUSIC without stereophonic supervision. Thus, it is a completely unsupervised setting. The idea is to show a rough prediction of separation results. During inference, the whole visual feature map is max-pooled to size $[6, 3, C_v]$ and average-pooled to $[2, 1, C_v]$. We then split them to two $[1, 1, C_V]$ feature maps according to their positions and rearrange them as $F_A$ and $F_B$ in equation (8) of the full paper. 

\end{itemize}

The results are shown in the second part of the video. We can see that while Mono2Binaural has only subtle influence under this setting, our Sep-Stereo can generate more plausible results according to the scenarios. As the distances between players in MUSIC are much smaller than that in FAIR-Play, it is reasonable to create less intense stereophonic audios. The results of our unsupervised model are over-intense as the model is borrowed from source separation. However, it is already a great advantage of our model to achieve stereophonic audio generation in an unsupervised manner.

\subsection{Audio Source Separation}

Finally, We show our source separation results qualitatively on MUSIC dataset. We mix the mono videos from the validation set as inputs. We explicitly compare our method with PixelPlayer~\cite{zhao2018sound}, which is designed only for separation.
It can be found that our results can outperform theirs for certain channels, which proves that we can achieve at least comparable results as PixelPlayer. 

\begin{figure*}[t]
    \includegraphics[width=1\linewidth]{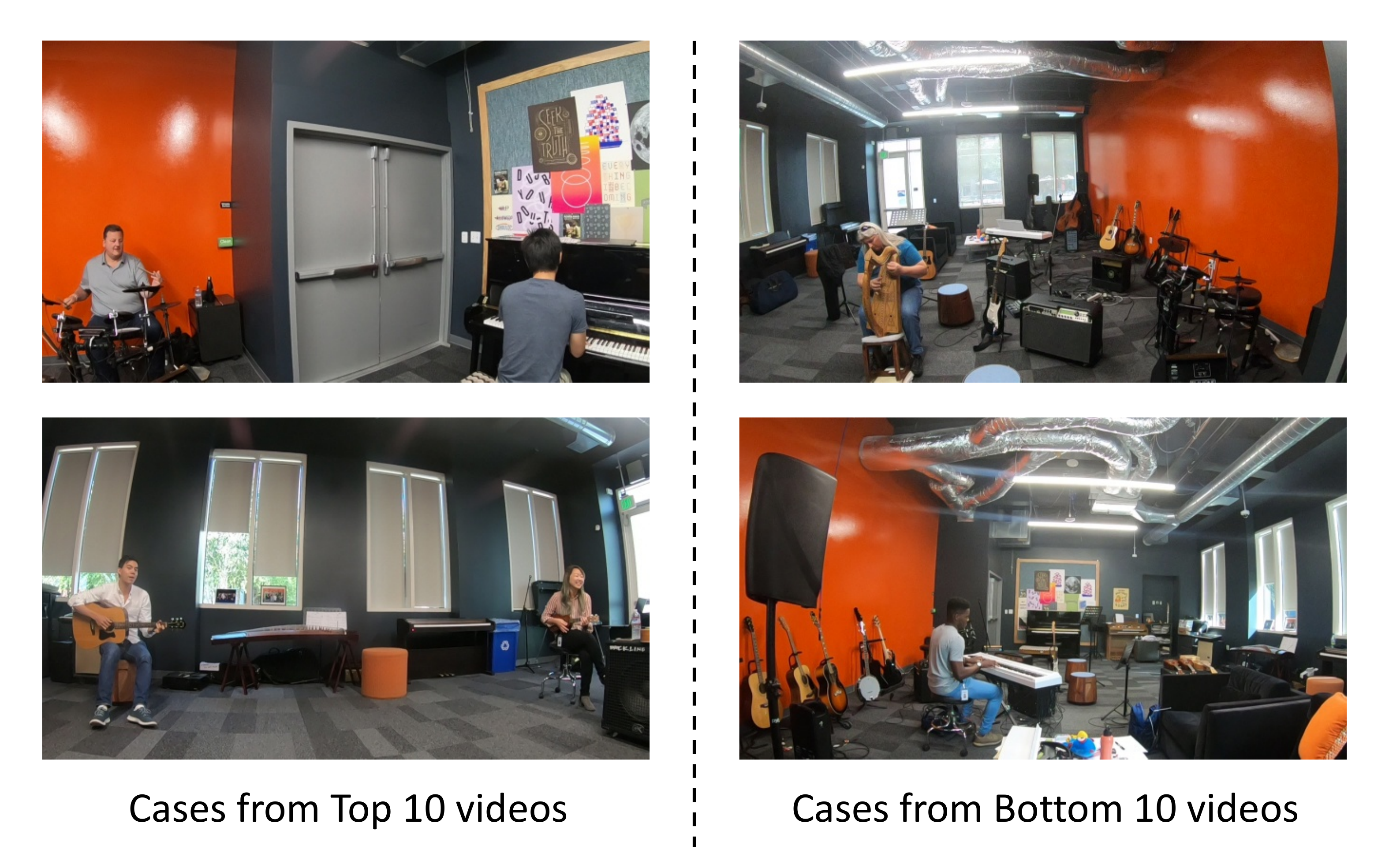}
    \caption{The cases from the top 10 and bottom 10 videos where our method outperforms Mono2Binaural~\cite{gao20192}}
\label{fig:result}
\end{figure*}


\section{Detailed Comparison with Mono2Binaural}

The differences between our method and that of Mono2Binaural~\cite{gao20192} are that 1) we propose separative learning, a completely new training diagram to leverage mono data for enriching stereo learning; and 2) we propose APNet for better visual-audio association in a more explainable way.

We compare our results with Mono2Binaural~\cite{gao20192}, both trained on the official split1 on FAIR-Play. The 187 results on the validation set are carefully examined. We rank the videos according to the performance gains between ours and~\cite{gao20192} w.r.t STFT differences. The larger the value, the better our result is.

1) Among the top 10 videos, we observe that the layouts are mostly with simple backgrounds. Sound sources are located at the far left and right sides of the view. This is reasonable and consistent with our designed framework with feature rearrangement. The results show that our separative learning indeed strengthened the learning of sound source horizontally.

2) Among the bottom 10 videos, the layouts are normally complicated with walls at two sides. The instruments are the rarely-appeared ones in both FAIR-Play and MUSIC. This also reveals the limitation of our work and can be improved in the future.

The arriving time and intensity difference of audios between left and right ears are the base of human sound localization ability. Thus the conclusion that our method can be more distinctive with horizontal cases is of great importance to stereo. Moreover, no previous work has shown results on both the tasks of separation and stereo simultaneously before.

\setlength{\tabcolsep}{8pt}
\begin{table}[t] 
\caption{More ablation results on FAIR-Play dataset}
\begin{center}

\begin{tabular}{ccccc}

\toprule

 Metric & w/o $L_D$& Vertical Extreme& Unet out & Ours
 \\
\noalign{\smallskip}
\midrule

$\text{STFT}_D$ &0.976 &0.919 & 0.933 & 0.879 \\

$\text{ENV}_D$ & 0.145 & 0.139 & 0.140 & 0.135 \\
\bottomrule
\label{table:ablation2}
\end{tabular}

\end{center}
\end{table}
\setlength{\tabcolsep}{1.4pt}

\section{More Ablations}

In this section, we provide more discussions together with ablation studies to make the design of our network clearer. The experiments are conducted on the FAIR-Play~\cite{gao20192} dataset. The results are shown in Table~\ref{table:ablation2}
\subsubsection{The loss $L_D$ at the output of the U-Net.} We found that the loss of $L_D$ is essential for stabling the training process. However, experiments show that the output results from the left and right branch independently from the APNet is better than the direct prediction (UNet out) of the difference map.
\subsubsection{The horizontal placement in feature rearrangement.} One may argue that we neglect the elevation differences in audios by placing the visual feature maps only at the left and right edges in separative learning, thus an experiment that places the feature maps at vertical extremes could be conducted. The discussions are as follows:

1) We choose only the horizontal extreme points based on the fact that human ears are distributed horizontally. Vertical information is difficult to represent with left-right channels.

2) Experiments with ``Vertical Extremes'' can be conducted by forming spatially inconsistent pairs (visual top to left channel and bottom to right), but the setting itself is somehow not reasonable. The results in the table~\ref{table:ablation2} show that the network benefits more from our normal layout.

3) However, the elevation issue is indeed a limitation of our work. Also, there is a domain gap between our self-created visual feature map and the directly extracted ones.

\end{document}